\documentclass[letterpaper, 10 pt, conference]{ieeeconf}

\IEEEoverridecommandlockouts                              
\usepackage{graphicx}
\usepackage{tabularx}
\usepackage[utf8]{inputenc}
\usepackage[T1]{fontenc}

\newsavebox{\twosubbox}
\usepackage{xcolor, hyperref}
\usepackage{color,soul}
\usepackage[ruled,vlined]{algorithm2e}
\usepackage{amsmath}
\usepackage{subcaption}
\usepackage{algpseudocode}
\usepackage{multirow}
\usepackage{booktabs}

\title{\LARGE \bf
Automatic Robot Path Planning for Visual Inspection from Object Shape
}

\author{Osama Tasneem$^{1}$ and  Roel Pieters$^{1}$
\thanks{$^{1}$Cognitive Robotics group, Unit of Automation Technology and Mechanical Engineering, Tampere University, 33720, Tampere, Finland; {\tt\small firstname.surname@tuni.fi}}%
}

\begin{document}

\maketitle
\thispagestyle{empty}
\pagestyle{empty}

\begin{abstract}

Visual inspection is a crucial yet time-consuming task across various industries. 
Numerous established methods employ machine learning in inspection tasks, necessitating specific training data that includes predefined inspection poses and training images essential for the training of models. The acquisition of such data and their integration into an inspection framework is challenging due to the variety in objects and scenes involved and due to additional bottlenecks caused by the manual collection of training data by humans, thereby hindering the automation of visual inspection across diverse domains.
This work proposes a solution for automatic path planning using a single depth camera mounted on a robot manipulator. Point clouds obtained from the depth images are processed and filtered to extract object profiles and transformed to inspection target paths for the robot end-effector. The approach relies on the geometry of the object and generates an inspection path that follows the shape normal to the surface. Depending on the object size and shape, inspection paths can be defined as single or multi-path plans. 
Results are demonstrated in both simulated and real-world environments, yielding promising inspection paths for objects with varying sizes and shapes.


Code and video are open-source available at: \textit{\footnotesize \url{https://github.com/CuriousLad1000/Auto-Path-Planner}}

\end{abstract}

\section{Introduction}\label{sec:intro}
\begin{figure}[t]
  \centering
  \includegraphics[width=0.5\textwidth]{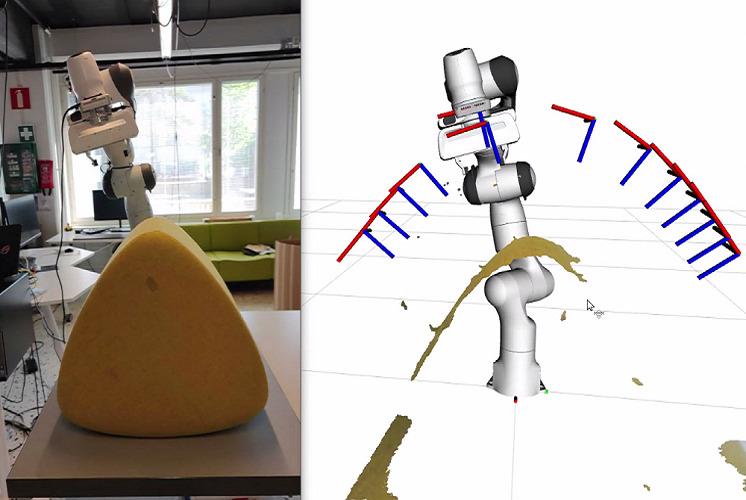}
  \caption{Automated path planning utilizes a robot with eye-in-hand camera to extract an inspection path that follows the geometry of the object.}
    \label{fig:overall}
\end{figure}

Visual inspection is an indispensable part of any industry \cite{See2017TheRO}. From product quality control to the maintenance of machines, visual inspection plays a vital part in improving the overall production yield and reducing the operation and maintenance costs. 
Due to the complexity of products and machines, the process of visual inspection is generally done by humans,
causing downtime and delay in production and, at the same time, exposing the person to hazardous working environments \cite{AGNISARMAN201952}.
However, in many cases, manual visual inspection is difficult to achieve, due to issues in accessibility or the high costs involved, creating demand for automated robotic solutions (see Fig. \ref{fig:overall}). 
For example, in the paper mill industry, suction rolls play a vital part in the production of paper from pulp. Based on the stage of use, they perform multiple functions including transportation of the sheet of paper, removing water from the paper, and dewatering the felt. This process can result in the blockage of the hundreds of thousands of holes in the suction roll. Current practices involve manual inspection of these rollers, which is time-consuming and ergonomically unfriendly. As the holes are small and narrowly spaced, operators are required to inspect the holes closely with special lighting equipment. Current growth in technologies has led to new possibilities where the task of visual inspection can be automated with sufficient accuracy \cite{Kato2020} by utilizing machine learning techniques.

However, the introduction of automation in inspection tasks does not intend to completely replace the human but to reduce the laborious task assigned to humans and improve the overall efficiency and accuracy of the system \cite{See2017TheRO}. 
Techniques such as deep learning \cite{He2015DeepRL, Liu2015SSDSS, Ronneberger2015UNetCN} 
are able to recognize various forms of anomalies that a human inspector used to look for and the implementation of such methods has produced better performance than humans on various applications such as object recognition  \cite{Dodge2017ASA, Kheradpisheh2015DeepNC} and sketch search \cite{Yu2017SketchaNetAD}. 
The accuracy of machine learning techniques being used to detect object anomalies heavily rely on the quality of data for training and validation. Collection of such data can become more difficult due to the size, shape, and location of the object. This is especially true for the inspection of larger infrastructures such as buildings and tunnels. Current data collection methods involve placing multiple cameras at different angles around an object \cite{Ho2013MulticameraFL} or placing a single camera \cite{Ren2021StateOT} on top and using an array of mirrors around the object to obtain different views \cite{Ali2019ACF}. 
As products and machines used in factories come in different sizes and shapes, re-arrangement of vision sensors is required to correctly align and gather sample data for training and validation. This introduces another bottleneck in the process to automate visual inspection.

\newpage
This paper proposes an approach that utilizes a single RGB-D camera, mounted on a robot manipulator to aid in the automation of the inspection process (see Fig. \ref{fig:overall}). Visual observations of the geometry of the object and/or the scene then determine the path for the robot to follow to perform visual inspection. 

Summarizing, the contributions of this work are:

\begin{itemize}
    \item an approach to extract an object profile from depth images 
    \item the generation of a path plan from the object profile for visual inspection 
    \item evaluation of the approach with robot and objects of various shapes in simulated and real-world experiments
\end{itemize}
The paper is organized as follows. We introduce the paper in Section \ref{sec:intro}. A brief overview of the current developments in path planning is given in Section \ref{sec:related_work}, which describes the state of the art and their limitations. The proposed method that extracts an inspection path given a point cloud is described in Section \ref{sec:problem} and its results are reported in Section \ref{sec:results}. The results are discussed along with its limitations in Section  \ref{sec:discuss}. Section \ref{sec:conclusion} concludes the work.

\section{Related Work}\label{sec:related_work}
This section provides an overview of related work in context to robot perception and path planning. 


Magnus et al. \cite{Hanses2016HandguidingRA} introduced a methodology in which the operator manually guides the robot along a predefined trajectory, ensuring compliance with joint constraints. The hand-guiding technique might be suitable for motion around objects during the inspection of smaller objects. However, it is not feasible to inspect larger objects using such technique due to time constraints and precision requirements of the task. Lončarević et al. \cite{Loncarevic2021SpecifyingAO} proposed an approach in which they leveraged existing CAD models of the target object to generate inspection trajectories around it. The operator is required to select the correct CAD model with the desired point along the required inspection path. This approach proves advantageous when a CAD model of the target is readily accessible, however, for inspecting unknown objects, an alternative methodology is needed. One approach, as detailed by Roberts et al. \cite{Roberts2017SubmodularTO}, involves the utilization of a trajectory optimization model. The authors applied this model to a drone for the purpose of conducting aerial 3D scanning. The method generated drone trajectories to capture and recreate a 'high-fidelity 3D model' of large structures. 
Another approach presented by Monica et al. \cite{Monica2018SurfelBasedNB} centers on Next Best View (NBV) planning, leveraging surfel representations of the environment,  
in lieu of intricate ray casting operations. The results demonstrate that a score function based on surfels is not only more computationally efficient but also achieves comparable outcomes in terms of reconstruction quality and completeness. Naazare et al. \cite{Naazare2022OnlineNP} also proposed an online NBV planner for a mobile manipulator robot. The proposed system is designed to facilitate comprehensive exploration as well as user-oriented exploration of an environment, including the inspection of regions of interest. A weighted sum-based information gain function was used to tackle exploration challenges characterized by multiple objectives. While NBV planners excel in robustly exploring unknown environments, it's important to note that these algorithms can be computationally expensive. Conversely, Fan et al. \cite{Fan2016AutomatedVA} presented an automated solution for 3D object acquisition. Their system implemented adaptive view planning and accommodated objects of varying scales. To devise the optimal path for a 2-axis manipulator, they formulated the path planning algorithm as a Traveling Salesman Problem. The utilization of point clouds as a source of information appears to be a promising and effective approach. However, dense point clouds result in large data volumes and higher processing times. In their work, Arav et al. \cite{Arav2022ContentAwarePC} introduced a context-aware subsampling technique, which effectively preserved the high-resolution details of objects while eliminating data from less critical regions. Research conducted by Yu et al. \cite{Yu2021PointCM} introduces a point cloud modelling and slicing algorithm designed to address free-form surfaces, specifically for the application of spray painting with a robot manipulator. This algorithm not only preserved the edges within the point cloud but also identified the optimal slicing direction and optimized the movement speed of the spray gun. The spraying trajectory points were based on the cross-section contour points. Wang et al. \cite{Wang2015ANP} introduced a path planning algorithm for robotic spray painting that is based on point cloud slicing. The authors employed an adaptive approach to determine the direction of the slice plane, in conjunction with the intersection-projection joint segmentation method, to obtain the points required for constructing the spraying path.

In comparison to the related studies, our approach does not require a CAD model, but generates a path from point cloud data by selecting and combining point clouds with the most data through a majority vote mechanism. The combined point cloud is refined with a hidden point removal process and downsampled to predetermined resolution. Surface normals are then estimated using covariance analysis of Eigen vectors of samples within the local neighborhood of points. 
Following, a clustering procedure employs DBSCAN clustering \cite{Ester1996ADA} to segment and extract valuable objects, after which a visual inspection path is planned.

\begin{figure*}[htp]
\centering
{
  \includegraphics[width=\linewidth]{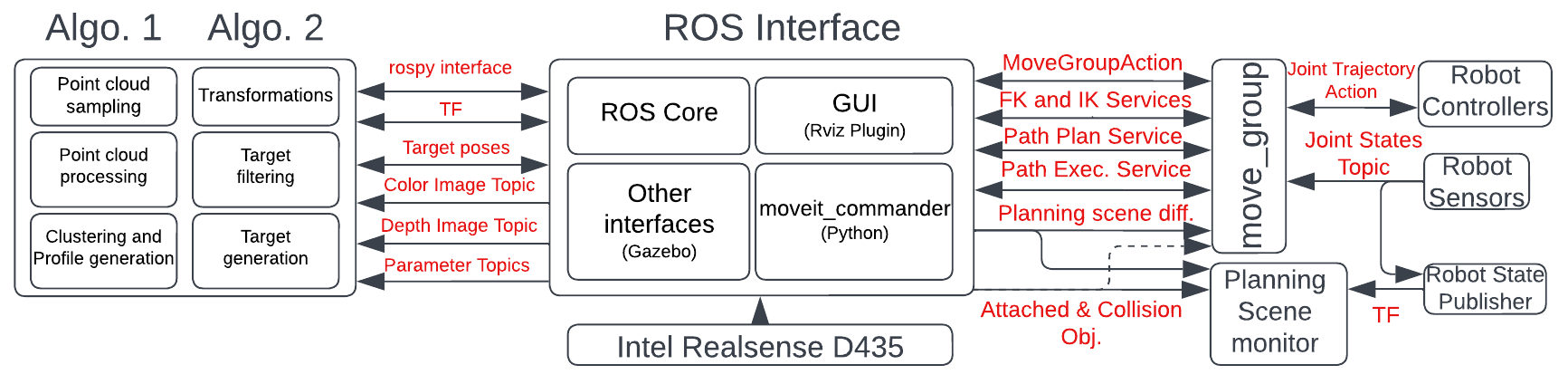} 
}\hfill
\caption{Block diagram depicting the flow of information between various modules.\label{fig:Block_diag}
}
\end{figure*}

\section{Methods}\label{sec:problem}
This section discusses the overall process in detail, with Alg. \ref{alg:one} and \ref{alg:two} as main methods. Alg. \ref{alg:one} describes the steps from the acquisition of a point cloud to the generation of its profile. Then, Alg. \ref{alg:two} describes the generation of robot target poses from this point cloud profile. Fig. \ref{fig:Block_diag} depicts a block diagram that highlights the information flow between various hardware and software blocks used and gives a general graphical overview of the system.


\vspace{5px}
\begin{algorithm}
\caption{Point cloud processing, clustering, and profile generation.}\label{alg:one}
\SetKwInOut{Input}{Input}  
\SetKwInOut{Output}{Output}
\SetKwInOut{Ensure}{Ensure}
\SetKwInOut{KwParams}{Parameters}
\KwParams{\\ 
        s: number of point clouds to sample\\
        $I_{RGB}$: RGB image frame from camera\\
        $I_D$: Depth image frame from camera\\
        $PCD$: container to hold point cloud\\
        $\mathbf{PCD}_a[]$: array for all sampled PCD\\
        $PCD_p$: Final object PCD profile\\
}

\Input{colour and depth frame, s}
\Output{Filtered point cloud profile $PCD_p$}
\textbf{STEP 1:}  Select initial pose of the robot.\\
\textbf{STEP 2:}  Object is in FoV of camera.\\

\textbf{STEP 3:}  Sample point clouds.\\
\ForEach{s}{
    $I_{RGB}$ , $I_D$ $\leftarrow$ grabframe()\\
    $PCD$ $\leftarrow$ GeneratePointCloud($I_{RGB}$, $I_D$)\\
    $PCD$ $\leftarrow$ FilterPCD($PCD$)\\
    $\mathbf{PCD}_a[]$ $\leftarrow$ $PCD$\\
}

\textbf{STEP 4:}  Select point clouds with a majority of similar number of points.\\
$\mathbf{idx_{list}}[]$ $\leftarrow$ SelectedPointClouds($\mathbf{PCD}_a[]$)\\

\ForEach{$\mathbf{idx_{list}}[]$}{
     $PCD$ $\leftarrow$  $PCD$ + $\mathbf{PCD}_a[idx]$\\
}

\textbf{STEP 5:}  Process point cloud and estimate normals.\\
$PCD$ $\leftarrow$ HiddenPointRemoval($PCD$)\\
$PCD$ $\leftarrow$ VoxelDownSample($PCD$)\\
$PCD$ $\leftarrow$ EstimateNormals($PCD$)\\

\textbf{STEP 6:}  Generate clusters and object profile.\\
$Clusters$ $\leftarrow$ ClusterPCD($PCD$)\\
$PCD$ $\leftarrow$ ClusterSelection($Clusters$)\\
$PCD_p$ $\leftarrow$ CroppedPCD($PCD$)\\

\end{algorithm}

\vspace{5px}
\begin{algorithm}
\caption{Target generation based on filtered point cloud profile.}\label{alg:two}
\SetKwInOut{Input}{Input}  
\SetKwInOut{Output}{Output}
\SetKwInOut{Ensure}{Ensure}
\SetKwInOut{KwParams}{Parameters}
\KwParams{\\ 
        $\mathbf{n}_p$: Normalized normals in ${PCD}_p$\\
        $\mathbf{u}_z$: Unit vector z-axis \\
        $\mathbf{n}_z$: Direction vec. by normal and z-axis\\
        $\theta_z$: Angle between normal and z-axis\\
        $\mathbf{R}_{\theta}$: Rot. matrix by normals and z-axis\\
        $\mathbf{T}_{p}$: Poses wrt object for ${PCD}_p$\\
        $\mathbf{T}_{e}$: EE poses wrt robot's base\\
}

\Input{Filtered Point cloud profile ${PCD}_p$}
\Output{EE Targets $\mathbf{T}_{e}$ wrt to robot's base}
\textbf{STEP 1:}  Generate Rotation matrix\\
$\mathbf{u}_z$ $\leftarrow$ $[0,0,1]$\\

\ForEach{${PCD}_p$}{
    $\mathbf{n}_p$ $\leftarrow$ normalize(${PCD}_p.normals$)\\
    $\mathbf{n}_z$ $\leftarrow$ cross($\mathbf{n}_p$, $\mathbf{u}_z$)/normalize(cross($\mathbf{n}_p$, $\mathbf{u}_z$))\\
    $\theta_z$  $\leftarrow$ -(acos(dot($\mathbf{n}_p$, $\mathbf{u}_z$)))\\
    $\mathbf{R}_{\theta}$ $\leftarrow$ RotMAxisAngle($\theta_z$$\mathbf{n}_z$)\\
    $\mathbf{T}_{p}$ $\leftarrow$  CalcTransforms($\mathbf{R}_{\theta}$,${PCD}_p.points$)\\
}
\textbf{STEP 2:}  Filter world coordinates\\
$\mathbf{T}_{p}$ $\leftarrow$ FilterAnomalies($\mathbf{T}_{p}$)\\
$\mathbf{T}_{p}$ $\leftarrow$ FilterThreshold($\mathbf{T}_{p}$)\\
$\mathbf{T}_{p}$ $\leftarrow$ FilterCloseCoords($\mathbf{T}_{p}$)\\

\textbf{STEP 3:}  Reorder the coordinates\\
$\mathbf{T}_{p}$ $\leftarrow$ Reverse($\mathbf{T}_{p}$)\\

\textbf{STEP 4:}  Generate EE targets by transformations\\
$\mathbf{T}_{e}$ $\leftarrow$ GenerateEETargets($\mathbf{T}_{p}$)\\

\end{algorithm}

\subsection{Point cloud acquisition and sampling}\label{sample}
Data is collected via a depth camera in the form of a point cloud. 
To account for noise and enhance robustness, we introduce a sampling filter (step 3 in Alg. \ref{alg:one}) that captures multiple frames of color and depth images ($I_{RGB}$, $I_D$) to generate multiple point clouds, as stored in $\mathbf{PCD}_a[]$ array. These multiple point clouds are then combined to create a better distributed and more dense version of the point cloud data ($PCD$).

\subsection{Point cloud processing}
The acquired point cloud $PCD$ undergoes processing to eliminate undesired elements, including the ground, any visible portions of the robot within the frame, and any other unwanted objects (step 3 in Alg. \ref{alg:one}). This filtering procedure is iteratively applied over a predetermined set of sample frames. Subsequently, the optimal frame is identified through a majority vote mechanism, mitigating outliers that deviate from the majority in terms of point count (step 4 in Alg. \ref{alg:one}).  Background points are eliminated using hidden point removal \cite{Zhou2018Open3DAM}. Afterward, the point cloud undergoes downsampling to a specified resolution. The importance of downsampling becomes particularly evident when managing dense point clouds, as they can strain computational resources, potentially obscure objects of interest \cite{Arav2022ContentAwarePC, Zhou2018Open3DAM} and are unnecessary for our specific application.

\subsection{Normal estimation}
The points within the point cloud contain data about the object's surface in a three-dimensional space relative to the depth camera. In essence, each point can be transformed to obtain its precise position in the world. However, achieving high-quality data for visual inspection necessitates the camera being positioned orthogonal to the points. These normals are estimated using covariance analysis of Eigen vectors of samples within the local neighborhood of points, a technique described in \cite{Pauly2003PointPF} and implemented through Open3D \cite{Zhou2018Open3DAM}.


\subsection{Clustering}
Due to the distribution characteristics of a captured point cloud, it is essential to cluster and delineate distinct groups of points within the point cloud and isolate an object of interest. Consequently, the DBSCAN clustering algorithm \cite{Ester1996ADA} is utilized, which 
provides the user with a GUI populated with various clusters of which the most useful ones are to be selected. Based on the cluster(s) selected, an object point cloud profile is generated. This profile is then passed on to Alg. \ref{alg:two} to generate targets and stored in $PCD_p$.

\subsection{Target generation}
The process of generating targets is defined by Alg. \ref{alg:two} and takes the object profile $PCD_p$ as input. While the points within the point cloud profile can be used to determine the position of each point, achieving the manipulator's desired pose also requires generating accurate orientations. To achieve this, the algorithm iterates over the points in the profile and computes a rotation matrix $\mathbf{R}_{\theta}$ (step 1 of Alg. \ref{alg:two}) by calculating the axis-angle representation for each point. This rotation matrix is then used to calculate the correct transformations for the object profile $PCD_p$ in world coordinates $\mathbf{T}_{p}$. Three separate filters are applied to the coordinates (step 2 in Alg. \ref{alg:two}) to filter out targets that are beyond user-defined thresholds and anomalous in nature. The first filter assesses coordinate order for each x, y, and z axis, checks for increasing or decreasing series by comparing each coordinate to the middle number and filters out coordinates if there's more than one axis with inconsistent trends. The second filter removes all coordinates that are too close to the ground or other surfaces and may pose danger to the movement of the end-effector. The third filter calculates the Euclidean distance between each point and if the distance is less than the 2 times the downsample resolution of the point cloud, the coordinate is filtered out.
The list of generated targets $\mathbf{T}_{p}$ is arranged in the correct sequence (step 3 in Alg. \ref{alg:two}), and precise transformations are applied to derive the end-effector's world coordinates $\mathbf{T}_{e}$ for each of these targets (step 4 in Alg. \ref{alg:two}).

\section{Results}\label{sec:results}

\subsection{Integration}\label{sec:integration}
For robot and perception hardware, we utilize the Franka Emika collaborative robot\footnote{\url{https://franka.de/}} and an Intel Realsense D435 camera, mounted on the end-effector of the robot for eye-in-hand perception. All developments are integrated with ROS\footnote{\url{https://www.ros.org/}\label{ROS}} within Jupyter\footnote{\url{https://jupyter.org/}} notebooks using Python. Additionally, Open3D \cite{Zhou2018Open3DAM} is used for 3D data processing of camera images and MoveIt!\footnote{\url{https://moveit.picknik.ai/}\label{fn2}} for the robot motion planning. The OMPL \cite{sucan2012the-open-motion-planning-library} motion planner provided by MoveIt! was selected as a default planner. The MoveIt python API provides the interface to the task space controller that is primarily used for the robot manipulator's motion. The waypoints are published to the MoveIt! API to make the robot move and record the images and video of the object. Computations are done on a Ubuntu PC with Nvidia GTX 1060 GPU, running ROS Noetic. The Gazebo simulation tool serves as the platform for creating a simulated environment in which our virtual objects are situated and where the experiment is conducted. The primary algorithms, along with the graphical user interface (GUI), is implemented using the Python programming language and runs within a Jupyter notebook. 
For testing, complex shaped objects with different curvatures and size were selected in both simulated and real environments. The objects larger than the robot manipulator were tested in the simulator. These include simulated models of car, tunnel, satellite, submarine, sphere and wall. Smaller models were less than half a meter in length and included physical objects such as bench, aerofoil, 3D printed model of a suction roll, textured metal sheets, inclined planes, etc.

\subsection{Point cloud sampling results}

To evaluate the effectiveness of the point cloud sampling technique described in Section \ref{sample}, we performed the following experiments. We positioned a real inclined metal sheet within the field of view of the depth camera. Adjacent to the metal sheet, we placed a strobe light oriented partially toward the camera. Throughout the experiments, the strobe light operated at a frequency of 3Hz with a 50 percent duty cycle. This setup simulated the varying ambient light intensity conditions affecting both the camera and the surrounding objects. The experiments were divided into three scenarios with varying point cloud samples $s \in\{1,5,10\}$, each of which was repeated five times to enhance precision. 

\begin{figure*}[htp]
\centering
\subcaptionbox{\label{fig:c1}}{%
  \includegraphics[height=0.22\linewidth]{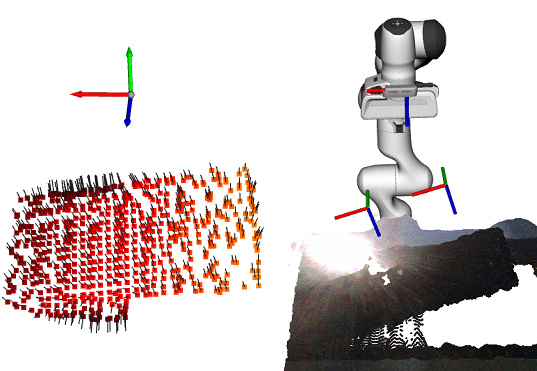}
}\hfill
\subcaptionbox{\label{fig:c2}}{%
  \includegraphics[height=0.22\linewidth]{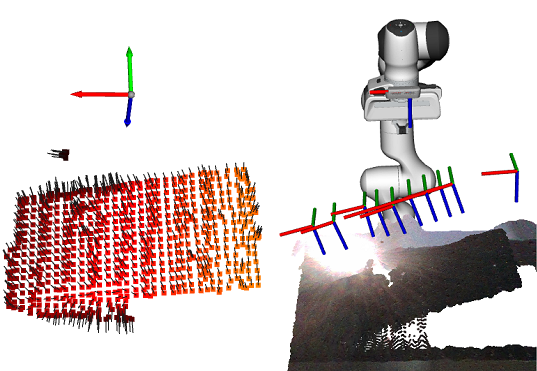} 
}\hfill
\subcaptionbox{\label{fig:c3}}{%
  \includegraphics[height=0.22\linewidth]{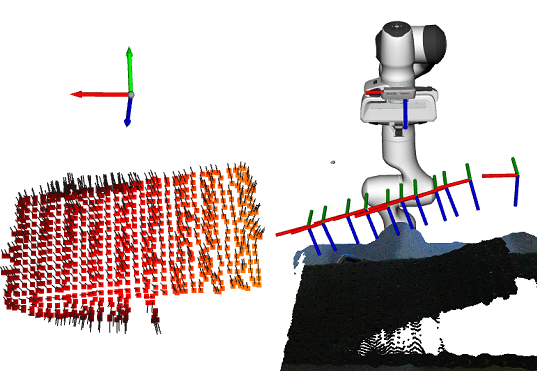} 
}\hfill
\caption{Point cloud sampling (step 3 in Alg. \ref{alg:one}) results of an inclined plane for $s=1$ with 629 points (a), $s=5$ with 792 points (b) and $s=10$ with 812 points (c). Sampling provides a better distribution of points and a higher density.
\label{fig:result_sampling1_filter}
}
\end{figure*}

The resulting point cloud with a single sample $s=1$ exhibited noticeable gaps in data across the surface and produced sub-optimal targets that did not cover the entire length of the object (see Fig. \ref{fig:c1}). In contrast, the combined point cloud was more reliable with sampling set to $s=5$ and $s=10$, as shown in Fig. \ref{fig:c2} and Fig. \ref{fig:c3} respectively. Ultimately, the results, shown in Table \ref{table:T1}, indicate that a higher number of samples leads to a better distribution of points and a higher density within the point cloud, which improved the total number of final targets generated. A compromise, however, is the increase in sampling time, with more samples included in a point cloud. Determining the ideal number of samples is dependent on current ambient conditions. There can be cases where the filter functions better with ten or more samples.

\begin{table}[t]
\centering
\renewcommand{\arraystretch}{1.5}
\caption{Point cloud sampling results (mean of five trials) for varying number of samples $s$.
}
\label{tab:T1}
\begin{tabular}{|l|c|c|c|c|}
\hline
 &
  \begin{tabular}[c]{@{}c@{}}Number of \\ points\end{tabular} &
  \begin{tabular}[c]{@{}c@{}}Sampling\\ time {[}sec{]}\end{tabular} &
  \begin{tabular}[c]{@{}c@{}}Object profile \\ points\end{tabular} &
  \begin{tabular}[c]{@{}c@{}}Final targets \\ generated\end{tabular} \\ \hline
$s=1$  & 715 & 0.48 & 25 & 8  \\ \hline
$s=5$  & 795 & 1.35 & 28 & 11 \\ \hline
$s=10$ & 791 & 2.03 & 28 & 10 \\\hline
\end{tabular}
\label{table:T1}
\end{table}


\subsection{Point cloud slicing results}
\begin{figure*}[htp]
\centering
\subcaptionbox{\label{fig:a1}}{%
  \includegraphics[height=0.245\linewidth]{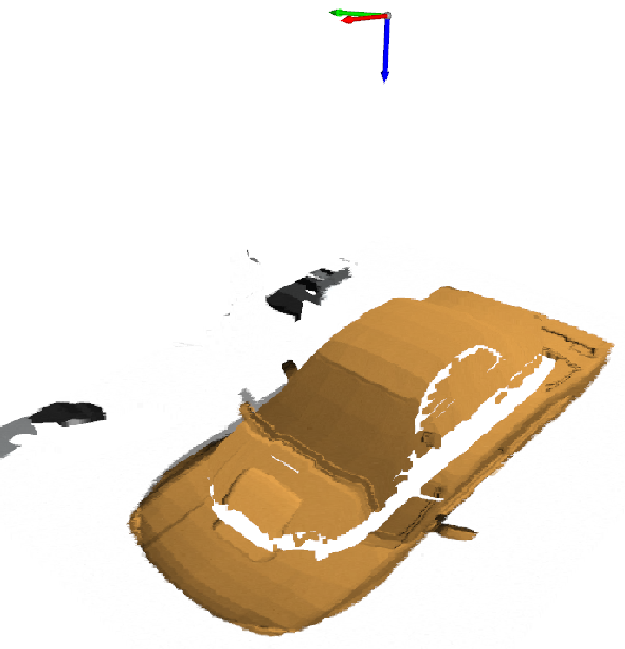} 
}\hfill 
\subcaptionbox{\label{fig:a2}}{%
  \includegraphics[height=0.245\linewidth]{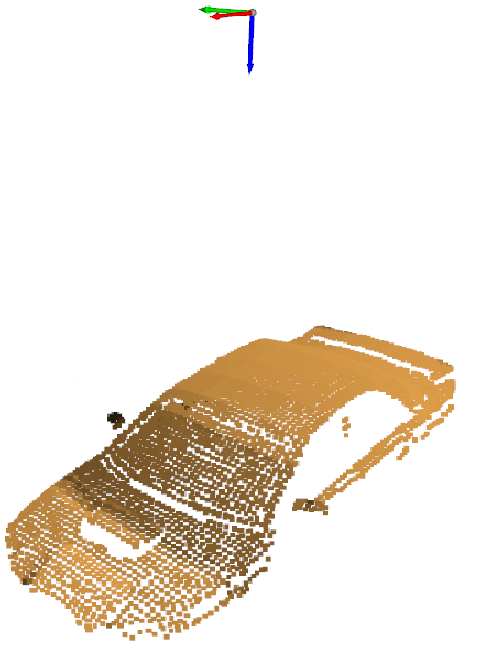}
}\hfill 
\subcaptionbox{\label{fig:a3}}{%
  \includegraphics[height=0.245\linewidth]{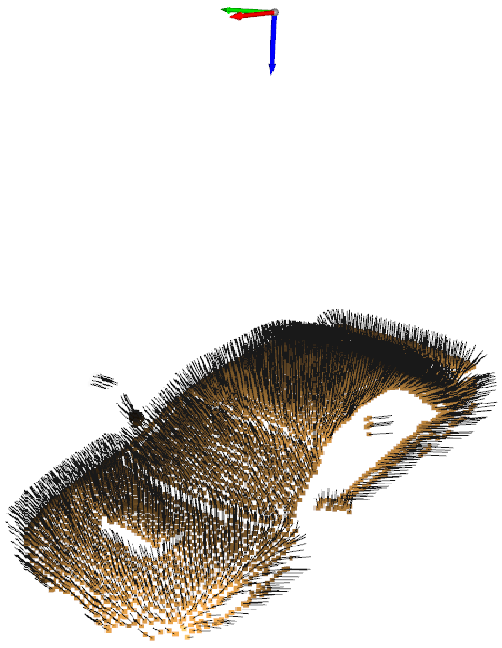} 
}\hfill 
\subcaptionbox{\label{fig:a4}}{%
  \includegraphics[height=0.245\linewidth]{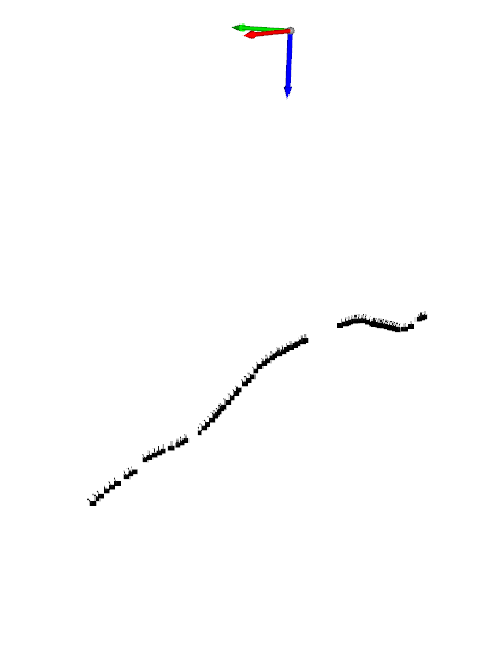} 
}\hfill 
\caption{Simulation results of point cloud filtering and profile generation. (a) depicts the raw point cloud from the depth camera. (b) shows the filtered and down-sampled point cloud (step 3 and 4 in Alg. \ref{alg:one}). (c) shows the outcome of the normal estimation process (step 5 in Alg. \ref{alg:one}). (d) shows the output generated by the profile generation algorithm (step 6 in Alg. \ref{alg:one}). \label{fig:result_PCD_filter}
}
\end{figure*}

Fig. \ref{fig:result_PCD_filter} demonstrates the results obtained after applying point cloud processing to a point cloud captured from a simulated object. In Fig. \ref{fig:a1}, the raw point cloud is depicted, containing both the primary object of interest and extraneous segments. After filtering and downsizing the point cloud (step 5 in Alg. \ref{alg:one}), Fig. \ref{fig:a2} is obtained and, following, normal estimation results are depicted in Fig. \ref{fig:a3}. The generated object profile (step 6 in Alg. \ref{alg:one}) is depicted in Fig. \ref{fig:a4}, from which the final targets are generated. 

\subsection{Path planning results}
Fig. \ref{fig:result_PCD_target}a-c depict the inspection paths and/or targets generated for a variety of simulated objects, including a down-scaled 3D model of a car with 28 targets, an aerofoil with 15 targets and a substantial 3D model of a tunnel with 106 targets. Fig. \ref{fig:result_PCD_target}d-g depict the inspection paths and/or targets generated using real-world objects, including a bench with 12 targets, an inclined metal sheet with 9 targets, an aerofoil with 13 targets and a model of suction roll with 4 targets. 
Target poses are generated based on certain physical constraints and can be adapted according to user preferences (see Step 2 in Alg. \ref{alg:two}), including the distance of targets to the surface and the spacing in between targets, which, in our experiments, was set to 0.3 meters and 0.02 meters, respectively.


\subsection{Multi-path planning results}
\begin{figure*}[htp]
\centering
\subcaptionbox{\label{fig:b1}}{%
  \includegraphics[height=0.245\linewidth]{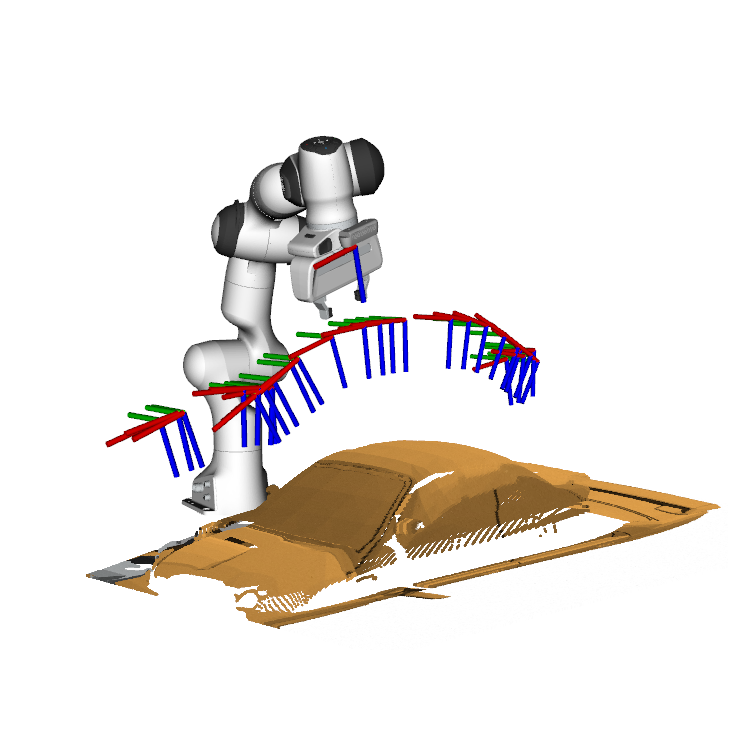} 
}\hfill 
\subcaptionbox{\label{fig:b2}}{%
  \includegraphics[height=0.245\linewidth]{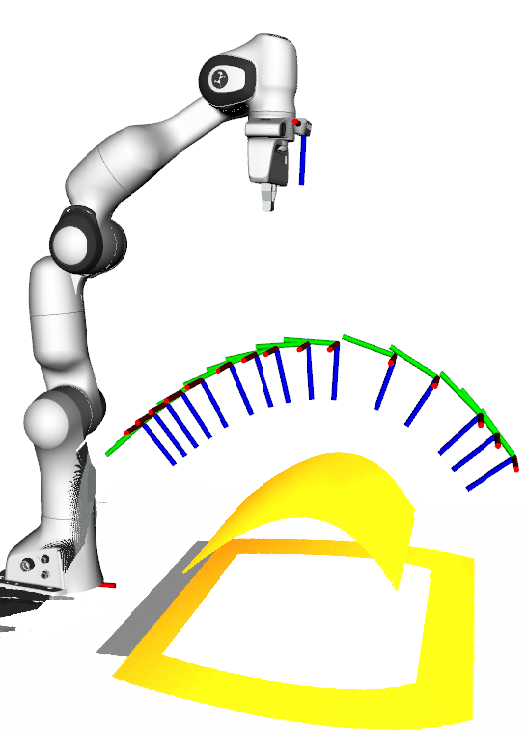} 
}\hfill 
\subcaptionbox{\label{fig:b3}}{%
  \includegraphics[height=0.245\linewidth]{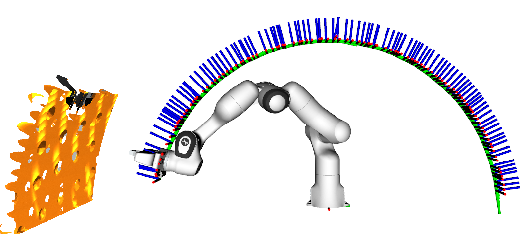} 
}\hfill 
\subcaptionbox{\label{fig:b4}}{%
  \includegraphics[height=0.245\linewidth]{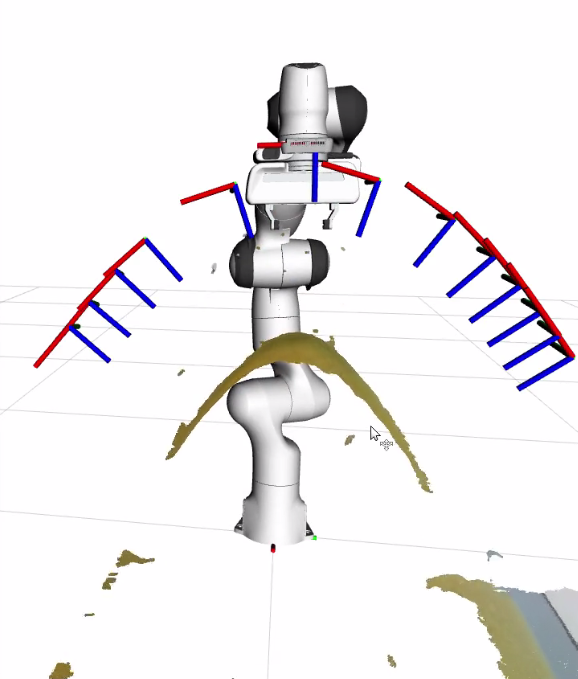}
}\hfill 
\subcaptionbox{\label{fig:b5}}{%
  \includegraphics[height=0.245\linewidth]{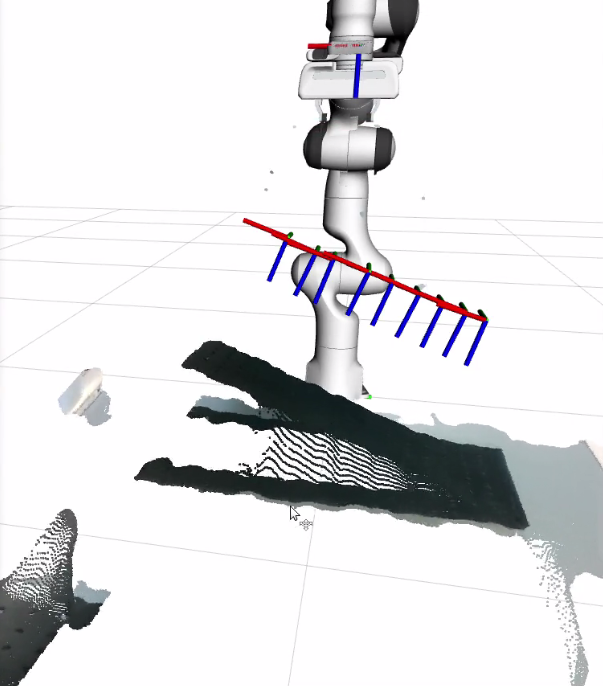} 
}\hfill 
\subcaptionbox{\label{fig:b6}}{%
  \includegraphics[height=0.245\linewidth]{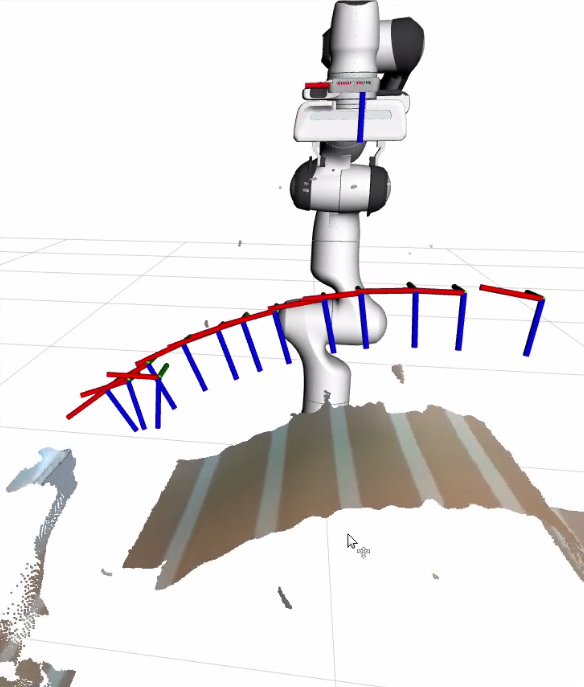} 
}\hfill 
\subcaptionbox{\label{fig:b7}}{%
  \includegraphics[height=0.245\linewidth]{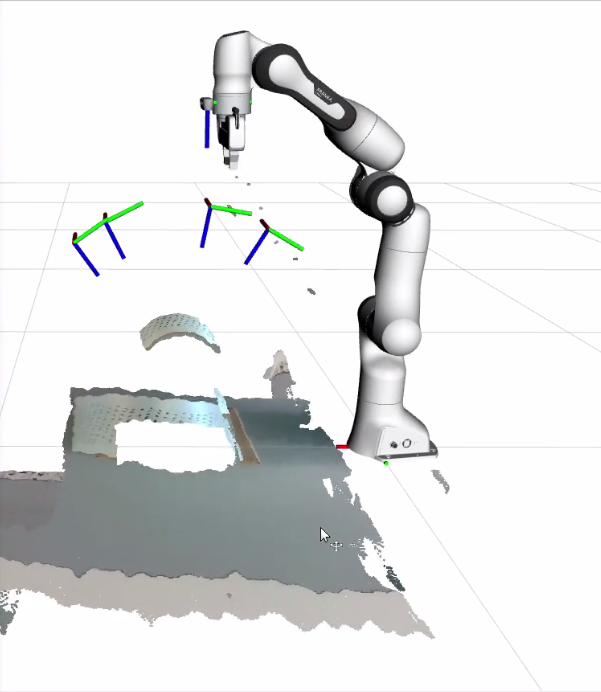} 
}\hfill 
\caption{Results of path targets generated for simulated (a-c) and real (d-g) objects. (a) depicts 28 targets generated across the surface of a simulated model of a car, (b) shows 15 targets generated following the shape of simulated model of an aerofoil, (c) depicts 106 targets generated across the cross section of a simulated model of a tunnel, (d) depicts 12 targets generated across a real bench, (e) depicts 9 targets generated across a real inclined metal sheet, (f) depicts 13 targets generated across the surface of a physical model of aerofoil, (g) depicts 4 targets generated for a 3D printed model of a suction roll.\label{fig:result_PCD_target}
}
\vspace{2mm}
\centering
\subcaptionbox{\label{fig:b8}}{%
  \includegraphics[height=0.245\linewidth]{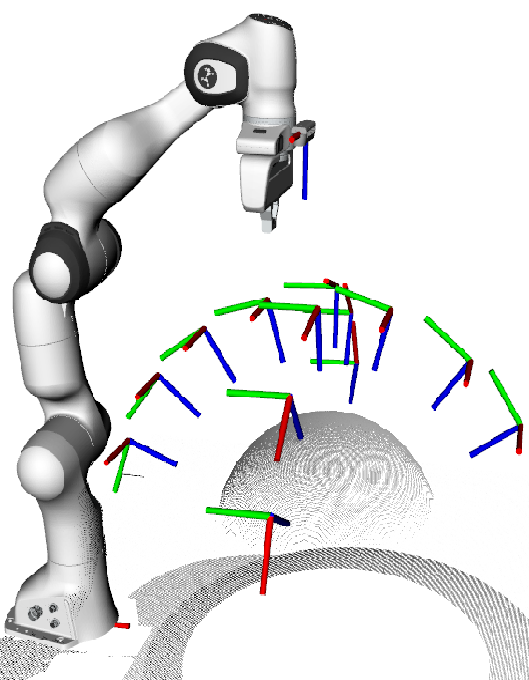}
}\hfill 
\subcaptionbox{\label{fig:b9}}{%
  \includegraphics[height=0.245\linewidth]{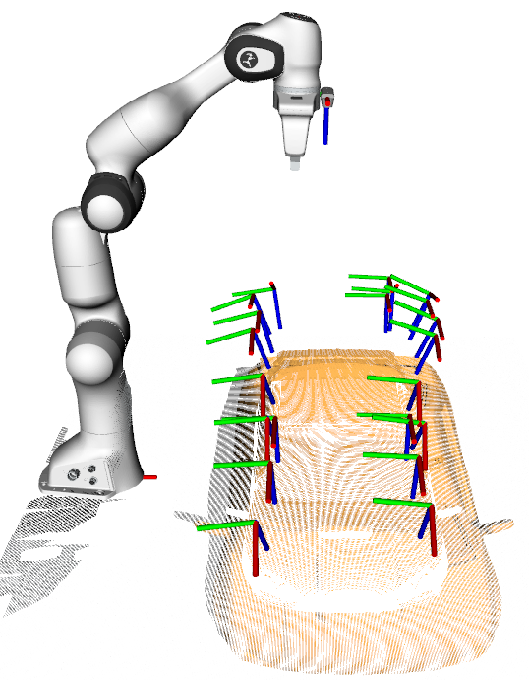} 
}\hfill 
\subcaptionbox{\label{fig:b10}}{%
  \includegraphics[height=0.245\linewidth]{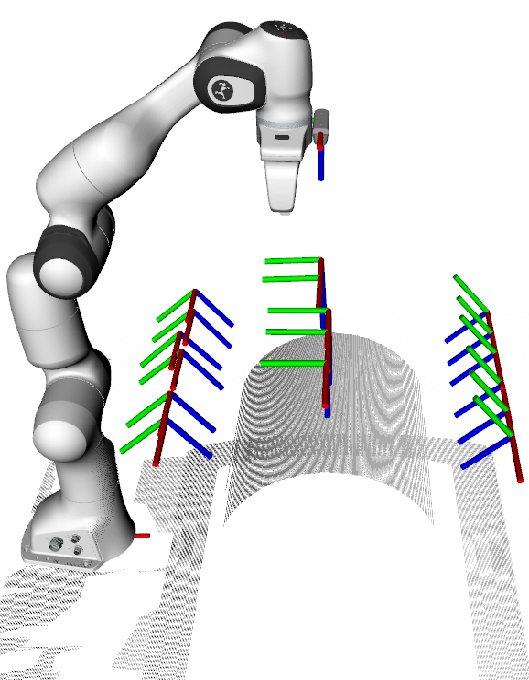} 
}\hfill 
\subcaptionbox{\label{fig:b11}}{%
  \includegraphics[height=0.245\linewidth]{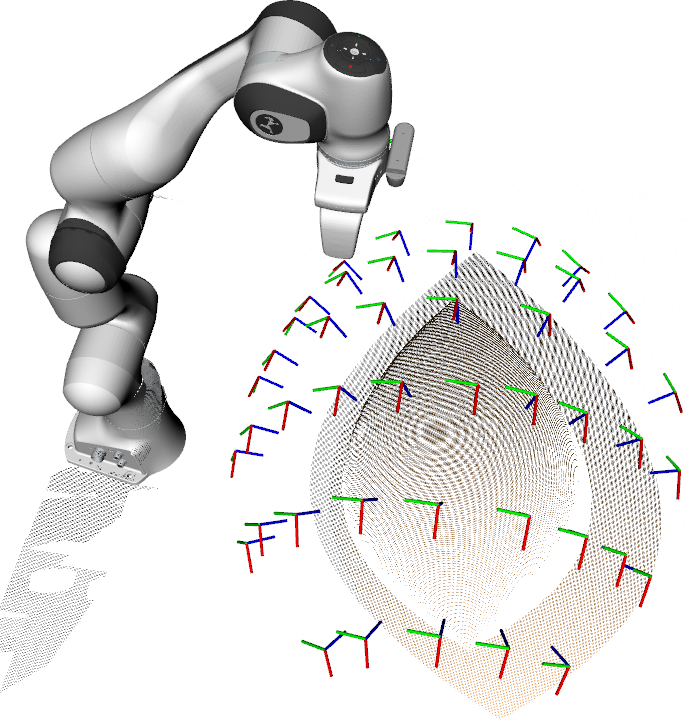} 
}\hfill 
\caption{Results of multi-path targets generated on simulated objects. (a) depicts 13 targets generated in two perpendicular directions for a spherical object. (b) depicts 21 targets generated in two rows covering both sides of a car, (c) depicts 17 targets generated in three rows covering the surface of a cylindrical object, (d) depicts 45 targets generated covering the entire surface of an irregularly shaped object. \label{fig:result_PCD_target_multi_path}
}
\end{figure*}

The previous results demonstrated the generation of paths with a single direction, i.e., along a single curvature of the object. Our approach also enables to select the direction for generating the object profile in accordance with a user's preferences. This means two distinct methods for determining the direction for generating an object profile. It can either automatically compute the direction based on the object's curvature, or it allows the user to specify a particular direction for profile generation. Additionally, users can select a specific segment to serve as the object profile. This effectively enables users to create numerous sequences of targets, offering comprehensive coverage of an object's surface, resulting in the generation of multi-path plans, as illustrated in Fig.  \ref{fig:result_PCD_target_multi_path}.

\section{Discussion}\label{sec:discuss}
As a general result, the path planning approach can be utilized for the visual inspection and 3D modelling of objects without any prior needed information such as object models, training data or human specified trajectories. Depending on the sampling rate of the point cloud, the approach can generate an inspection path in relatively short time with standard perception and computational hardware (see Section \ref{sec:integration}). In addition, our developments are robot-agnostic and are provided open-source to research community. 

\subsection{Simulated vs. real experiments}
Experiments included both simulated and real-world environments. In simulation, path plans demonstrated better performance due to the absence of physical factors like noise and ambient conditions. In contrast, conducting experiments in the real world proved to be more challenging due to these physical factors. After evaluating the results from both settings, additional filters (step 2 in Alg. \ref{alg:two}) were incorporated to enhance the resilience against noise and environmental variations. The integration of these filters resulted in improved outcomes in real-world scenarios, albeit at the expense of increased algorithm complexity and execution time.
Nevertheless, the approach offers users the flexibility to configure and optimize the filters to suit the specific characteristics of the object under inspection. 
\vspace{2mm}

The process of generating targets is influenced by three distinct parameters employed at various program stages. As the targets are derived from the processed point cloud, their characteristics are directly influenced by the resolution of the downsampled point cloud. In our experiments, the point cloud downsampling was configured to 0.02 meters, which implies that initial targets cannot be positioned closer than 2 centimeters unless the user further reduces the downsampling scale. The second stage involves the elimination of irregularities in the targets derived from the object profile. It also filters out targets that fall below a user-defined threshold and those that are positioned closer than twice the downsampling scale. Depending on the specific task at hand, this process filters out nearly 50 percent of the total generated targets. The final target filter is a straightforward decimation filter designed to reduce the number of targets by $n$ between two targets.

\vspace{2mm}
A final observation concerning simulation and real-world experiments relates to motion planning. As a separate motion planning and control system is responsible for determining the robot's trajectory to reach the inspection path targets, the choice of motion planner dictates whether the robot follows an optimal trajectory between these targets or not. Also, motion planning needs to consider the workspace and range of the robot, and avoid singular configurations, which is again affected by the choice of motion planner.

\newpage
\subsection{Comparison to related work}
The field of path planning around objects has been thoroughly explored by numerous researchers in diverse contexts. Some studies concentrate solely on determining the next best view for the purpose of exploring unfamiliar environments, as seen in \cite{Naazare2022OnlineNP} and \cite{Monica2018SurfelBasedNB}. In contrast, others focus on known objects and environments, by requiring CAD models as part of the path planning approach \cite{Loncarevic2021SpecifyingAO}. In our research, path planning aims to generate inspection targets around unknown objects, without considering score functions for next best views.


Concerning hardware utilized, certain studies employ drones, as indicated in \cite{Roberts2017SubmodularTO}, while others utilize mobile manipulator robots, as demonstrated by \cite{Naazare2022OnlineNP}. Conversely, some focus solely on perception, as shown in \cite{Wang2015ANP}, for the application of spray painting. In contrast, our research integrates the visual perception approach with a collaborative robot manipulator. It should be noted that our work is robot-agnostic and that any robotic system could utilize the inspection path planning approach.


\subsection{Limitations}
The primary objective in developing our approach was to create a path planner that could be applied universally to objects of varying shapes. While successful results are demonstrated in Section \ref{sec:results}, different limitations are observed as well.

\begin{figure}[!thp]
\centering
\subcaptionbox{\label{fig:out1}}{%
  \includegraphics[width=0.48\linewidth]{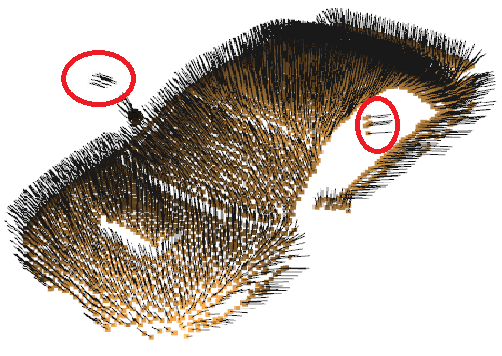} 
}\hfill
\subcaptionbox{\label{fig:out2}}{%
  \includegraphics[width=0.48\linewidth]{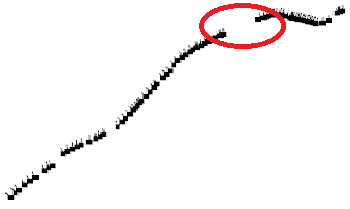}
}\hfill
\subcaptionbox{\label{fig:out3}}{%
  \includegraphics[width=0.5\linewidth]{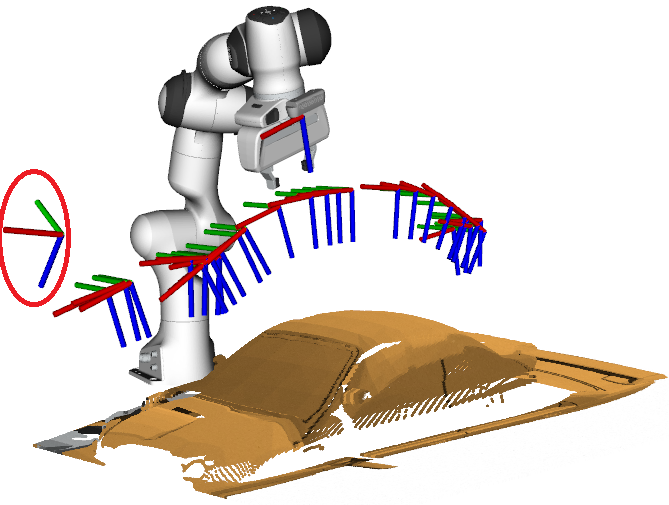}
}\hfill
\caption{Limitations on the approach include targets in erroneous locations (a), missing targets (b) and ghost targets (c). 
\label{fig:outlier_tgt}
}
\end{figure}

One observation during experiments pertained to the introduction of stray points within a point cloud. Given the intricate and dynamic shapes of the objects being inspected, certain cases still exhibited residual points in a point cloud despite using a comprehensive filtering process. For example, in Fig. \ref{fig:out1}, highlighted in red, the side window pane of the car and points on the left were not entirely filtered, leaving a small cluster of points. This may not pose an issue when generating the object profile from a position within the object's interior, but it could potentially lead to undesired targets when profiling from the object's edges or from a different side.

Furthermore, excessive filtering could have a detrimental effect, causing the profile to lose important details such as curves and edges of the object. In Fig. \ref{fig:out2}, highlighted in red, there appears to be a minor section absent from the car's roof within the object profile. This absence should not pose significant implications for the inspection task, as long as that part remains within the camera's field of view while moving towards the next target. 

Finally, despite incorporating multiple filters within the point cloud processing, there remains the possibility of encountering unexpected outlier points. These outliers have the potential to destabilize the system, making it imperative to fine-tune predefined parameters to mitigate such issues. For example, during the target generation phase, sporadic 'ghost' targets were observed, appearing out of order and in undesirable locations. An example of this phenomenon is visible in Fig. \ref{fig:out3}, where the initial target materialized unexpectedly and without any discernible point of origin. Our approach includes internal mechanisms to attempt the reordering and filtering of these targets, however, there may be instances where the filtering process doesn't function as intended. The appearance of these ghost targets appeared to be entirely random, with a suspected cause related to communication timing between the algorithms and ROS. While our approach offers functionalities to filter targets based on collision thresholds, users are advised to exercise caution and preview generated targets within Rviz before executing any actions.

\vspace{25px}


\section{CONCLUSION}\label{sec:conclusion}
In this paper, we proposed an automatic path-planning solution for visual inspection, based on the object shape. This addresses the problem of generating a path (position and orientation) that traverses objects of different curvatures and scales, from single or multiple instances of point clouds. 
The point cloud undergoes a filtering and clustering process to extract the object profile, depending on few constraints (user-defined thresholds, number of point cloud samples, clustering parameters) set by the user. Both single paths and multi-paths can be generated according to the size of the objects to inspect.
The approach was evaluated using both simulated and real-world objects of varying sizes and shapes, demonstrating successful path generation and motion execution of a robot with eye-in-hand camera. As a results, the approach can be utilized for the visual inspection and 3D modelling of objects without any prior needed information such as object CAD models or training data. In addition, the approach can run on standard hardware, with single RGB-D camera and any desired robot hardware.  

\section*{Acknowledgements}
This project has received funding from the European Union's Horizon 2020 research and innovation programme under grant agreement no. 871252 (METRICS). 

\bibliographystyle{IEEEtran}
\bibliography{IEEEabrv,refs}

\end{document}